\pdfoutput=1

\documentclass[11pt]{article}
\usepackage[final]{acl}
\usepackage{times}
\usepackage{latexsym}
\usepackage[T1]{fontenc}
\usepackage[utf8]{inputenc}
\usepackage{microtype}
\usepackage{inconsolata}
\usepackage{graphicx}
\usepackage{xspace}
\usepackage{tabularx}
\usepackage{booktabs}
\usepackage{caption}
\usepackage{multirow}
\usepackage{makecell}
\usepackage{caption}
\usepackage{subcaption}
\usepackage{enumitem}
\usepackage{amssymb}

\usepackage{todonotes}

\newcommand{\myframework}{\textsc{DVAGen}\xspace}
\newcommand{\dvamodel}{\textsf{DVAModel}\xspace}
\newcommand{\dvatokenizer}{\textsf{DVATokenizer}\xspace}
\newcommand{\phrasesampler}{\textsf{PhraseSampler}\xspace}
\newcommand{\retriever}{\textsf{Retriever}\xspace}
\newcommand{\dvalogitsprocessor}{\textsf{DVALogitsProcessor}\xspace}

\newcommand{\trainable}{\includegraphics[height=0.8em]{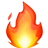}\xspace}
\newcommand{\frozen}{\includegraphics[height=0.8em]{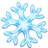}\xspace}

\title{\myframework: Dynamic Vocabulary Augmented Generation}

\author{
  Wei Du\textsuperscript{1}\thanks{\ Equal Contribution}\,,
  Nuowei Liu\textsuperscript{1}\footnotemark[1]\,,
  Jie Wang\textsuperscript{1}\footnotemark[1]\,,
  Jiahao Kuang\textsuperscript{1}\footnotemark[1]\,,\\
  \textbf{Tao Ji\textsuperscript{2},}
  \textbf{Xiaoling Wang\textsuperscript{1},}
  \textbf{Yuanbin Wu\textsuperscript{1}\thanks{\ Corresponding Author}} \\
  \textsuperscript{1}School of Computer Science and Technology, East China Normal University\\
  \textsuperscript{2}School of Computer Science, Fudan University \\
  \texttt{\{weidu.cs, nwliu, jiewang.cs, jhkuang\}@stu.ecnu.edu.cn}\\
  \texttt{taoji@fudan.edu.cn, \{xlwang, ybwu\}@cs.ecnu.edu.cn}
}

\begin{document}
\maketitle
\thispagestyle{plain}
\begin{abstract}
    Language models trained with a fixed vocabulary struggle to generalize to novel or out-of-vocabulary words, limiting their flexibility in handling diverse token combinations. Existing dynamic vocabulary approaches attempt to address this limitation but face challenges such as fragmented codebases, lack of support for modern LLMs, and limited inference scalability. To overcome these issues, we introduce \myframework, a fully open-source, unified framework designed for training, evaluation, and visualization of dynamic vocabulary-augmented language models. Our framework modularizes the pipeline for ease of customization, integrates seamlessly with open-source LLMs, and is the first to provide both CLI and WebUI tools for real-time result inspection. We validate the effectiveness of dynamic vocabulary methods on modern LLMs and demonstrate support for batch inference, significantly improving inference throughput.\footnote{Codes are publicly available at \url{https://github.com/AntNLP/DVAGen}.}
\end{abstract}

\section{Introduction}
\label{sec:intro}
Most language models \cite{grattafiori2024llama, yang2025qwen3} are typically trained on a fixed vocabulary, which constrains their ability to generalize beyond the training corpus. This limitation becomes apparent when encountering novel (out-of-vocabulary) words or when attempting to efficiently generate phrases composed of arbitrary token combinations.

Instead of relying solely on a pre-trained tokenizer with a static vocabulary, recent studies have adopted dynamic vocabulary approaches to augment the generation process, thereby enhancing modeling capabilities for natural languages \cite{lan2023copy,cao2024retrieval,liu2024generation} and even protein languages \cite{liu2025protein}.

However, these prior works rely on diverse codebases, some of which are not yet fully open sourced, and lack a unified framework for training, inference, and visualization of results augmented by dynamic vocabulary. Furthermore, they primarily focus on GPT-2 based language models \cite{radford2019language,ferruz2022protgpt2}, without validation on Large Language Models (LLMs). Additionally, the absence of support for batch inference further increases inference latency, despite the methods demonstrating promising inference speed when processing a single input at a time.

To address these issues, we present \myframework and our key contributions are highlighted below.

\textbf{A fully open-source, unified framework for training, evaluation, and visualization.}
We propose a unified framework that decouples individual modules, allowing users to customize each component. The framework supports the full training and inference pipeline (\verb|train|, \verb|chat|, and \verb|eval|) for developing dynamic vocabulary-augmented applications. Furthermore, we are the first to offer both command-line tools and a WebUI interface for visualizing generation results.

\textbf{Integration of existing open-source LLMs and their validation.}
Existing open-source LLMs can be easily integrated into \myframework in a plug-and-play manner. We further evaluate the language modeling performance of LLMs using our framework and validate the effectiveness of employing dynamic vocabulary methods on LLMs.

\textbf{Supports batch inference to enhance inference throughput.}
Dynamic vocabulary methods demonstrate promising capabilities in reducing token usage during decoding. Owing to their sequence compression abilities, these methods exhibit low inference latency. Building on this foundation, \myframework supports batch inference, further maximizing inference speed.



\section{Background}
\label{sec:background}

Research on dynamic vocabulary generation addresses key limitations of standard language models that rely on \textbf{static tokenization}. Traditional approaches like Byte-Pair Encoding (BPE) and WordPiece employ fixed vocabularies \citep{radford2019language,devlin2019bert}, which struggle to incorporate domain-specific phrases or multi-token expressions during generation. Efforts to enhance static vocabularies, such as Multi-Word Tokenization (MWT) \citep{gee2023multi}, expand vocabularies with frequent n-grams but remain inflexible post-training.

\textbf{Retrieval-augmented} methods have emerged as alternatives. Document-level approaches like RETRO \citep{borgeaud2022improving} integrate retrieved passages via cross-attention but incur high latency. The Copy-is-All-You-Need (CoG) framework \citep{lan2023copy} pioneered phrase-level retrieval through a two-stage pipeline: first retrieving relevant documents, then extracting candidate phrases. While CoG supports variable-length generation, it processes phrases as token sequences during input, creating inconsistency between input and output representations. CoG-2 \citep{cao2024retrieval} advanced this paradigm by enabling direct phrase retrieval without document pre-filtering. Using syntactic heuristics and self-reinforced training, CoG-2 improved knowledge-intensive task performance but maintains the output-only phrase handling limitation of its predecessor.

The \textbf{dynamic vocabulary} approach \citep{liu2024generation} constitutes a fundamental architectural shift. It introduces a causal Transformer-based phrase encoder that maps arbitrary text spans to atomic units, unifying their treatment in both input and output layers. This eliminates the token-phrase inconsistency seen in CoG variants. The method operates plug-and-play, requiring no model retraining when incorporating new phrases. The dynamic vocabulary’s end-to-end atomic processing provides a more unified solution for fluency, efficiency, and verifiability in text generation.

\section{\myframework}
\label{sec:method}
\begin{figure*}[!htb]
    \centering
    \includegraphics[width=\linewidth]{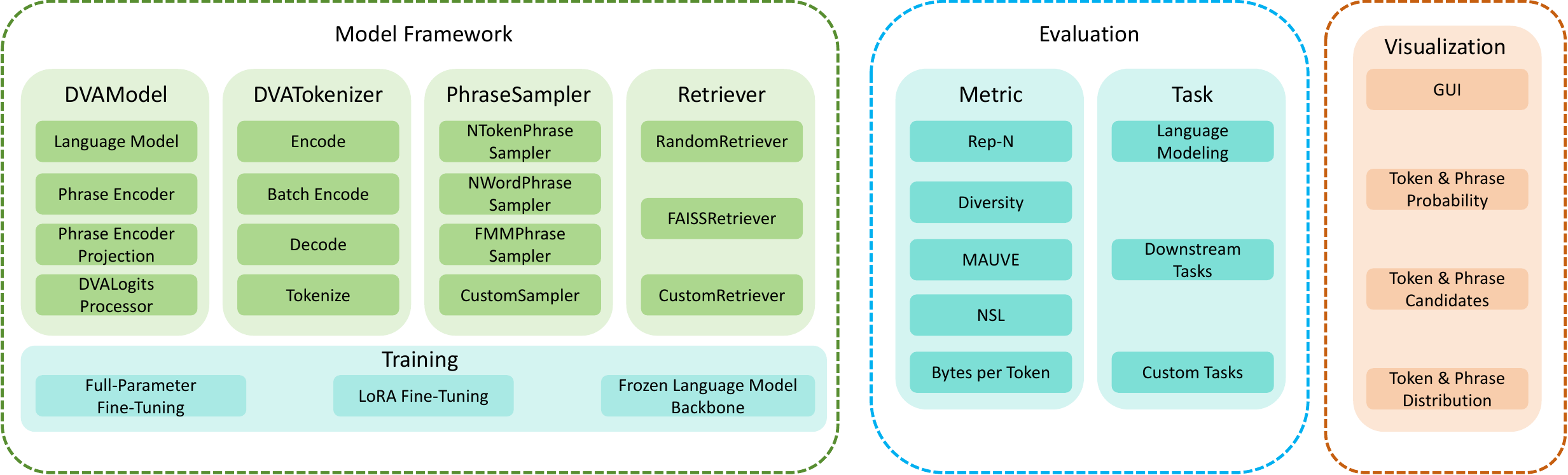}
    \caption{The overall architecture of \myframework. The light color boxes represent modules, while the dashed boxes indicate
collections of modules with relevant functions.}
    \label{fig:framework}
\end{figure*}

In this section, we introduce the overall architecture of \myframework, as illustrated in Figure~\ref{fig:framework}.

\subsection{DVAModel}

The DVA framework integrates a dynamic phrase encoder with a standard language model (LM) through a projection layer, enabling on-the-fly vocabulary expansion. Formally, given a static vocabulary $V$ and a phrase candidate set $P = \{p_1, \dots, p_m\}$ from \phrasesampler, the model operates as follows:

\paragraph{Phrase Encoding} For each phrase $p_i \in P$, tokenize it into subwords $[w_1, \dots, w_s]$ using the Phrase Encoder’s static tokenizer (e.g., GPT-2’s BPE), then compute its embedding via the Phrase Encoder:
$$
\mathbf{e}_{p_i} = \text{Projector}\left(\text{PhraseEncoder}(\mathbf{w}_{1:s})_s\right)
$$  
where $\mathbf{w}_{1:s}$ denotes subword embeddings of $p_i$, and the last token's hidden state is taken as the vector representation of $p_i$. A Projector then maps this to the LM’s embedding space. Typically, the Phrase Encoder is a causal Transformer model, and the Projector is an MLP.
    
\paragraph{Vocabulary Expansion} Concatenate phrase embeddings with LM's original input/output embedding matrices:
$$
\mathbf{E}_{\text{in}}' = \left[\mathbf{E}_{\text{in}}, \mathbf{E}_P\right], \quad 
\mathbf{E}_{\text{out}}' = \left[\mathbf{E}_{\text{out}}, \mathbf{E}_P\right]
$$  
where $\mathbf{E}_P = [\mathbf{e}_{p_1}, \dots, \mathbf{e}_{p_m}] \in \mathbb{R}^{d \times m}$. The expansion is dynamic, that is for each sentence, we can attach different $\mathbf{E}_P$ for generation.
\paragraph{Generation} The \dvamodel then generate on the expanded dynamic vocabulary. At step $t$, we compute logits on both token vocabulary $V$ and phrase candidates $P$, and choose the output $y_t \in V \cup P$ accordingly:
$$
y_t\sim \text{Softmax}(h_t\mathbf{E}_{\text{out}}')\in\mathbb{R}^{|V|+|P|}
$$
where $h_t$ is the last hidden state of the LM. We note that \myframework allows each sentence in a batch to have a different set of phrase candidates, in which case a phrase mask is incorporated in DVA Logits Processor to filter out a specific set in $\mathbf{E}_P$ for that sentence.

\subsection{PhraseSampler}

The \phrasesampler $\mathcal{S}$ extracts candidate phrases $P$ from retrieved documents $D$ using configurable strategies. \myframework provides a set of pre-defined strategies:

\begin{itemize}[leftmargin=1em]
    \item \textit{NToken}: Samples $n$-gram token sequences from tokenized text.
    \item \textit{NWord}: Extracts word-level $n$-grams via word tokenization (e.g., using NLTK).
    \item \textit{FMM} (Forward Maximum Matching): Suppose the current position in the text is $i$ from a corpus $\mathcal{C}$, then a segment $\mathcal{C}[i:i+m]$ is identified such that it appears in at least one sentence of the corpus $\mathcal{C}$, and $m$ is the maximum value satisfying this condition. This segment will be extracted as a phrase, and the process continues from position $i+m$.
\end{itemize}

Custom samplers can be integrated by implementing the interface $P \leftarrow \mathcal{S}(D, \text{config})$.

\subsection{DVATokenizer}
Denote input text as $X={x_{1}, x_{2},...,x_{L}}$ where $L$ is the length of input text. There are three main function in \dvatokenizer. 
\paragraph{Tokenize} This function can split text to phrases and tokens according to $P$:
$$
\operatorname{Tokenize}(X,P)=\{x_{1}', x_{2}',...,x_{L'}'\},L' \leq L
$$  
where $x_{i}'\in{P \cup X}$ is original token or candidate phrase.
\paragraph{Encode} It can encode input text to input ids.
$$
    \operatorname{Encode}(X,P)=\{\mathrm{id_1}, \mathrm{id_2},...,\mathrm{id}_{L'}\},L' \leq L
$$
$$
        \operatorname{type}(\mathrm{id}_{i}) = 
        \begin{cases} 
        \text{token}, & \text{if } \mathrm{id}_i < |V|  \\
        \text{phrase}, & \text{otherwise}
        \end{cases}
$$
If the $\mathrm{id}_{i}$'s value exceeds the vocabulary size $|V|$, it is a token ID; otherwise, it is a phrase ID.

\paragraph{Decode} This function is the inverse function of encode which can get the original text from the mixed ids.
$$
        \operatorname{Decode}(\operatorname{Encode(X,P)})=X
$$


\subsection{Training}
\label{sec:train}

\begin{figure}[!htb]
    \centering
    \includegraphics[width=\linewidth]{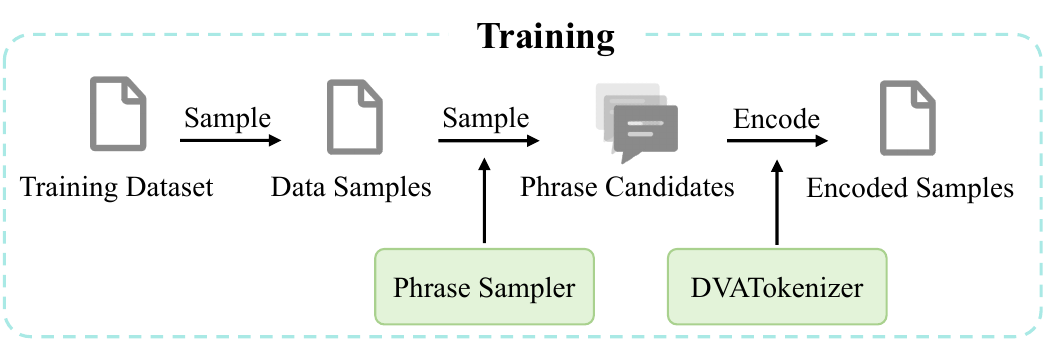}
    \caption{The overall training pipline of \myframework.}
    \label{fig:framework-training}
\end{figure}

\myframework supports full-parameter fine-tuning, LoRA fine-tuning \cite{hu2022lora}, and training with a frozen language model backbone. Different training strategies share a similar processing paradigm, as illustrated in Figure~\ref{fig:framework-training}. Given a training dataset \(D\), we first sample a batch containing \(m\) training samples. A \phrasesampler is then applied to sample phrases from the batch. With an arbitrary combination of tokens and phrases within each training sample, we employ a \dvatokenizer to encode them into discrete token and phrase ids for subsequent training.

\subsection{Inference and Evaluation}
\label{sec:infer}

\begin{figure}[!htb]
    \centering
    \includegraphics[width=\linewidth]{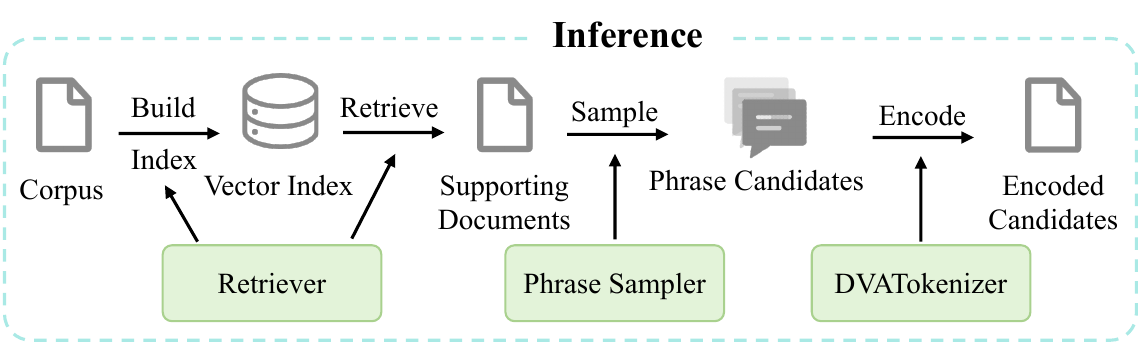}
    \caption{The overall inference pipline of \myframework.}
    \label{fig:framework-inference}
\end{figure}

\paragraph{Inference}
The inference paradigm differs from the training process by incorporating the retrieval of the top \(k\) most relevant supporting documents. The overall inference pipeline is illustrated in Figure~\ref{fig:framework-inference}. Given an input prefix, we employ a \retriever implemented using an embedding model with a FAISS backend \cite{douze2024faiss} to enable efficient similarity search on dense vectors. Building on top of LangChain\footnote{\url{https://github.com/langchain-ai/langchain}}, we utilize the GPU version of FAISS \cite{johnson2019billion} to accelerate the retrieval process. The retrieved supporting documents are then used to sample phrases using a \phrasesampler, and the sampled phrases serve as candidates for constructing the dynamic vocabulary. It is worth noting that for different input prefixes, the phrase candidates vary according to the retrieved supporting documents. Unlike previous studies \cite{lan2023copy, liu2024generation} that do not support batch inference, we implement the \dvalogitsprocessor to mask irrelevant logits (i.e., those corresponding to phrase ids associated with phrases unrelated to the current input). The positions to be masked are determined based on phrase candidates from other inputs within the same batch.

\paragraph{Evaluation}
Evaluation across various tasks are supported in \myframework, including basic language modeling and a range of downstream tasks. To ensure ease of use and fair comparisons, our framework includes implementations of a wide array of evaluation metrics. These metrics are categorized into three types: (1) generation quality (e.g., MAUVE \cite{pillutla2021mauve}, Rep-N and Diversity \cite{welleck2019neural}, and Perplexity), (2) sequence compression (e.g., Normalized Sequence Length and Bytes per Token \cite{dagan2024getting}), and (3) downstream task performance (e.g., Recall, Precision, F1, and ROUGE-L \cite{lin2004rouge}).

\subsection{Visualization}
Our web application is developed using the Django\footnote{\url{https://www.djangoproject.com/}}, featuring a front-end interface built with Bulma\footnote{\url{https://bulma.io/}} to provide a clean and responsive user experience, and \myframework handles all text generation tasks in backend. 

\begin{figure}[!htb]
    \centering
    \includegraphics[width=\linewidth]{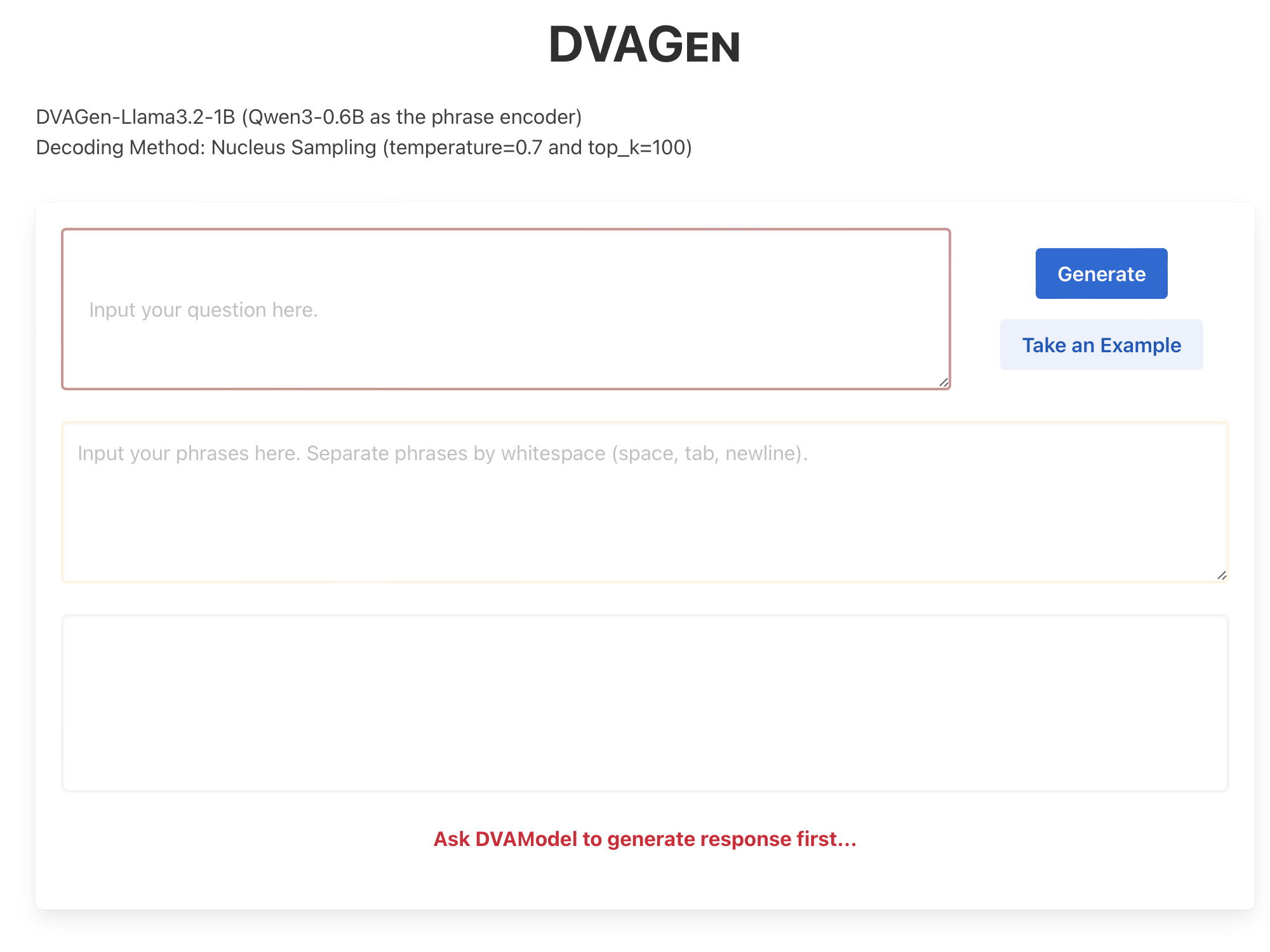}
    \caption{The screenshot of the \myframework WebUI interface.}
    \label{fig:vis-web-input}
\end{figure}

Figure~\ref{fig:vis-web-input} presents the input area of the front-end interface. The brown input box allows users to enter an initial query or prefix text, while the yellow phrase input area beneath it is specifically designed for defining dynamic vocabulary—users can specify phrases that the model should prioritize during generation. After completing the inputs, users can click the "Generate" button to initiate the generation process.

\begin{figure}[!htb]
    \centering
    \includegraphics[width=\linewidth]{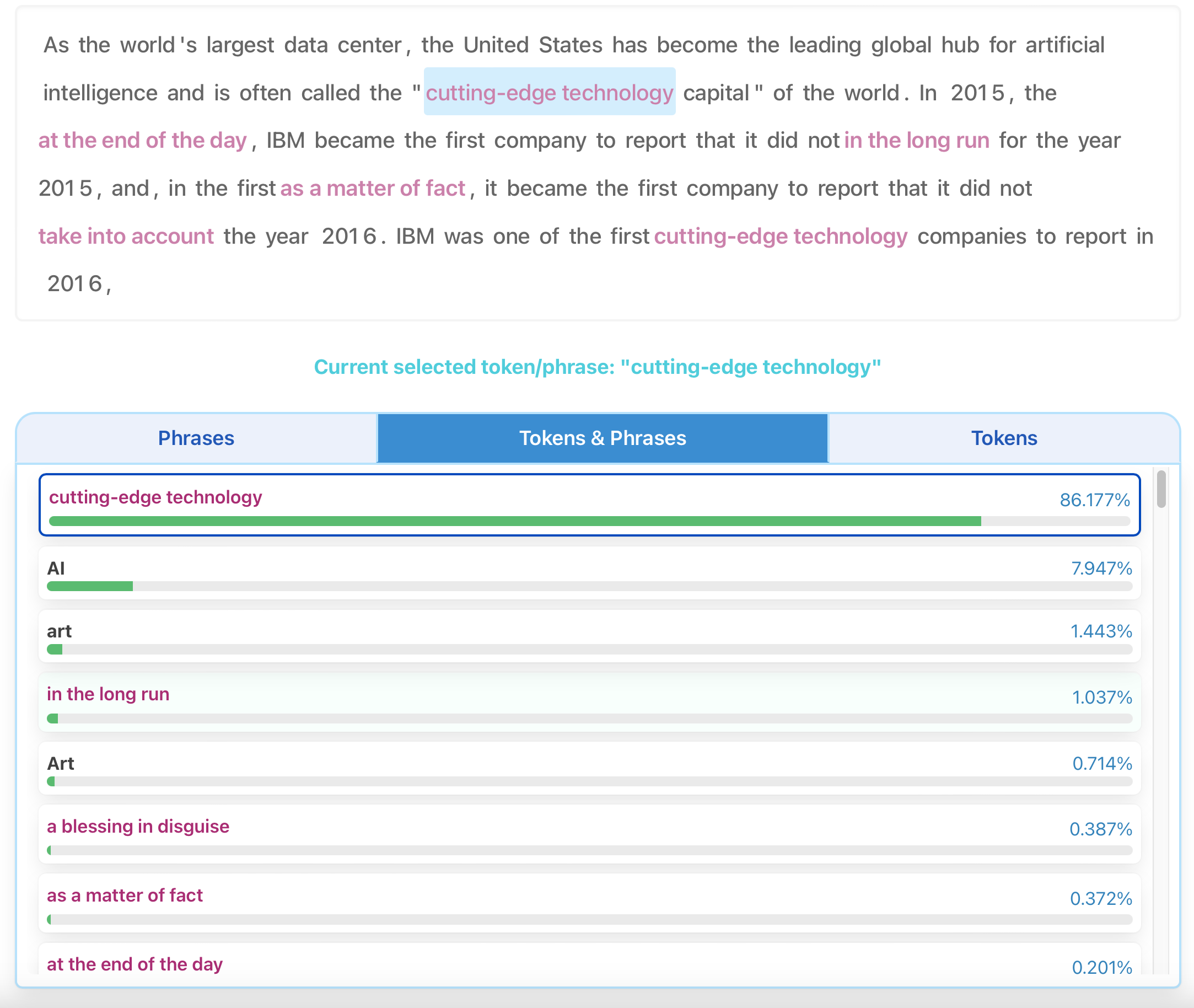}
    \caption{The screenshot displays the generation outputs. Phrases are highlighted in red, and the logit probabilities for each token and phrase can be viewed by clicking on the respective token or phrase.}
    \label{fig:vis-web-output}
\end{figure}

Figure~\ref{fig:vis-web-output} displays the output area of the interface after generation is complete. The generated text appears in the grey output box. By clicking on any token or phrase in the box, users can view a scrollable list of candidate alternatives for that position, along with their associated probabilities. Users may then select a preferred alternative to replace the original token or phrase, thereby constructing a new prefix for controlled generation. To support more targeted interaction, the interface also offers filter options such as "Phrases", "Tokens \& Phrases" and "Tokens", enabling users to refine the candidate list by category and focus on the specific categoriy of alternatives.

To intuitively present the complex outputs generated by our model, we have developed a powerful visualization application.

\begin{figure}[!htb]
    \centering
    \includegraphics[width=\linewidth]{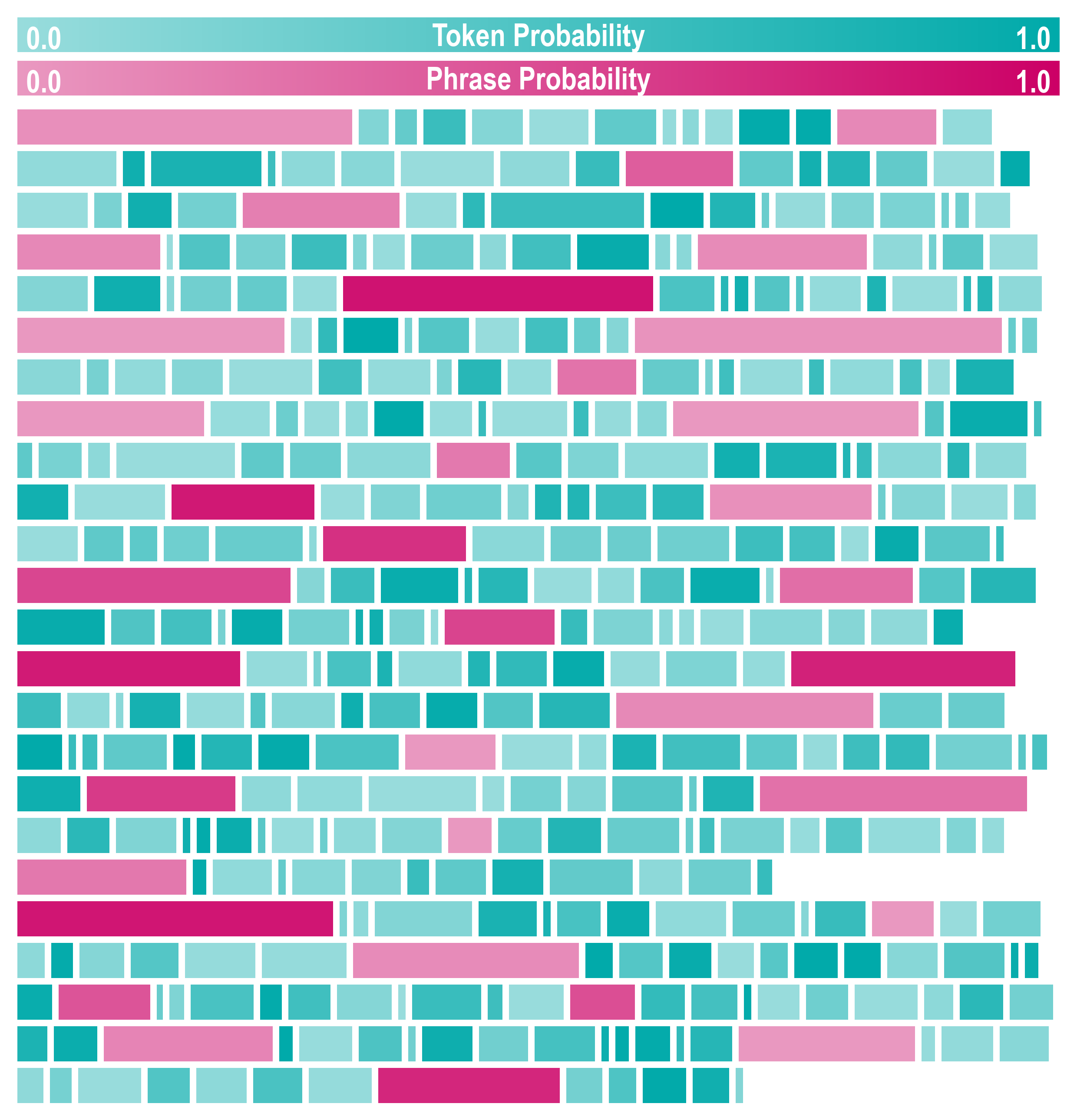}
    \caption{Visualization of generation outputs is represented in blue for token distribution and red for phrase distribution. The brightness of the colors reflects the logit probabilities, with darker colors indicating higher confidence.}
    \label{fig:vis-visualization}
\end{figure}

Figure~\ref{fig:vis-visualization} illustrates the visualization generated by the application, which accurately captures the model's interpretation of each token and phrase within the generated text. Tokens and phrases are distinguished using distinct color schemes—for instance, blue represents tokens, while purple-red denotes phrases. The shade of each color reflects the corresponding generation probability, with deeper shades indicating higher probabilities. A heatmap positioned at the top of the visualization clearly maps probability values to color intensities. Additionally, the application offers extensive customization options, enabling users to adjust token and phrase colors, font sizes, and image resolution to suit their specific analytical or presentation requirements.

\section{Evaluation}
\label{sec:eval}
\subsection{Language Modeling}
\label{sec:exp-lm}

\paragraph{Setup}
We first show that \myframework can enhance the basic language modeling capabilities of LLMs. To evaluate this, we select the pre-trained base models Qwen3-0.6B\footnote{\url{https://huggingface.co/Qwen/Qwen3-0.6B-Base}}, Qwen3-1.7B\footnote{\url{https://huggingface.co/Qwen/Qwen3-1.7B-Base}}, Llama3.2-1B\footnote{\url{https://huggingface.co/meta-llama/Llama-3.2-1B}}, and Llama3.2-3B\footnote{\url{https://huggingface.co/meta-llama/Llama-3.2-3B}}. These models are fine-tuned on the WikiText-103 training set to serve as baselines. To ensure a fair comparison across different model families, a GPT-2 \cite{radford2019language} tokenizer is employed to extract the first 32 tokens from each test sample as input. During inference with \myframework, we utilize the \textsf{NTokenPhraseSampler} to sample phrases from the given input prefixes. Qwen3-Embedding-0.6B\footnote{\url{https://huggingface.co/Qwen/Qwen3-Embedding-0.6B}} is used as the embedding model for \retriever, and 32 documents are retrieved for each sample.

\paragraph{Results}

As shown in Table~\ref{tab:lm-wikitext}, our key findings can be summarized as follows:

\textbf{(1) \myframework improves generation quality while requiring fewer tokens to generate the same content.}
\myframework helps maintain a lower Rep-N score and higher diversity, particularly for Llama models. The improvement in the MAUVE score indicates the generation of more natural and coherent language that closely matches human-written text. More importantly, training with \myframework significantly enhances sequence compression capability, demonstrating improved efficiency in reducing both tokens and decoding steps.

\textbf{(2) Freezing the language model during training is memory-efficient and achieves performance on par with that of a trainable backbone.}
The results highlight that training with a frozen language model backbone yields performance on par with a trainable backbone while significantly reducing memory requirements. The evaluations offer insights for future research aimed at developing memory-efficient methods and reducing model size to enable more efficient training and inference.

\begin{table*}[!htb]
    \centering
    \resizebox{\linewidth}{!}{
        \begin{tabular}{lcccccccc}
            \toprule
            Models & MAUVE \(\uparrow\) & Rep-2 \(\downarrow\) & Rep-3 \(\downarrow\) & Rep-4 \(\downarrow\) & Diversity \(\uparrow\) & PPL \(\downarrow\) & NSL \(\downarrow\) & \makecell{Bytes per\\Token \(\uparrow\)} \\
            \midrule
            Qwen3-0.6B & 21.70 & 11.46 & 5.07 & 3.05 & 81.49 & 51.09 & 1.00 & 3.17 \\
            \myframework-Qwen3-0.6B \frozen & 24.31 & 19.68 & 10.92 & 7.59 & 66.12 & 22.29 & 0.87 & 4.84 \\
            \myframework-Qwen3-0.6B \trainable & 24.41 & 14.81 & 6.46 & 3.61 & 76.82 & 24.19 & 0.85 & 4.89 \\ \midrule
            Qwen3-1.7B & 22.74 & 10.21 & 4.14 & 2.19 & 84.18 & 57.55 & 1.00 & 3.27 \\
            \myframework-Qwen3-1.7B \frozen & 24.92 & 18.64 & 10.26 & 7.21 & 67.76 & 21.73 & 0.86 & 4.88 \\ 
            \myframework-Qwen3-1.7B \trainable & 23.57 & 16.05 & 7.88 & 4.99 & 73.47 & 19.93 & 0.99 & 4.63 \\ \midrule
            Llama3.2-1B & 24.74 & 17.28 & 8.73 & 5.29 & 71.50 & 28.65 & 1.00 & 4.00 \\

            \myframework-Llama3.2-1B \trainable  & 24.64 & 15.70 & 7.41 & 4.46 & 74.57 & 81.66 & 0.88 & 4.90 \\ \midrule
            Llama3.2-3B & 25.61 & 15.62 & 7.64 & 4.58 & 74.35 & 63.04 & 1.00 & 4.01 \\
            \myframework-Llama3.2-3B \frozen  & 25.26 & 16.85 & 8.50 & 5.41 & 71.96 & 43.98 & 0.87 & 4.97 \\
            \myframework-Llama3.2-3B \trainable & 26.40 & 14.30 & 6.49 & 3.79 & 77.09 & 37.19 & 0.93 & 4.90 \\  
            \bottomrule
        \end{tabular}
    }
    \caption{Evaluation results on the WikiText-103 test set. Qwen3-0.6B is used as the phrase encoder and remains trainable across all settings. The \trainable and \frozen markers indicate whether the language model backbone is trainable or frozen, respectively.}
    \label{tab:lm-wikitext}
\end{table*}

\subsection{Inference Performance}
\label{sec:exp-inference}

\paragraph{Setup}
To evaluate the generation performance of \myframework, we compare the tokens and UTF-8 bytes generated per second by Qwen3-0.6B and \myframework-Qwen3-0.6B. For a fair comparison, both models are evaluated on the WikiText-103 test set under identical settings. Each sample in a batch consists of 32 tokens as input prefix, and the models are required to generate exactly 128 new tokens (or phrases when using \myframework) by setting both the minimum and maximum number of generated tokens to the same value. Note that, in alignment with previous work \cite{lan2023copy, liu2024generation}, the retrieval cost is not included, as the construction of phrase candidates can be conducted offline. The evaluations are conducted using a single NVIDIA RTX 4090 GPU.

\begin{figure}[!htb]
    \centering
    \begin{subfigure}[b]{0.48\linewidth}
        \centering
        \includegraphics[width=\linewidth]{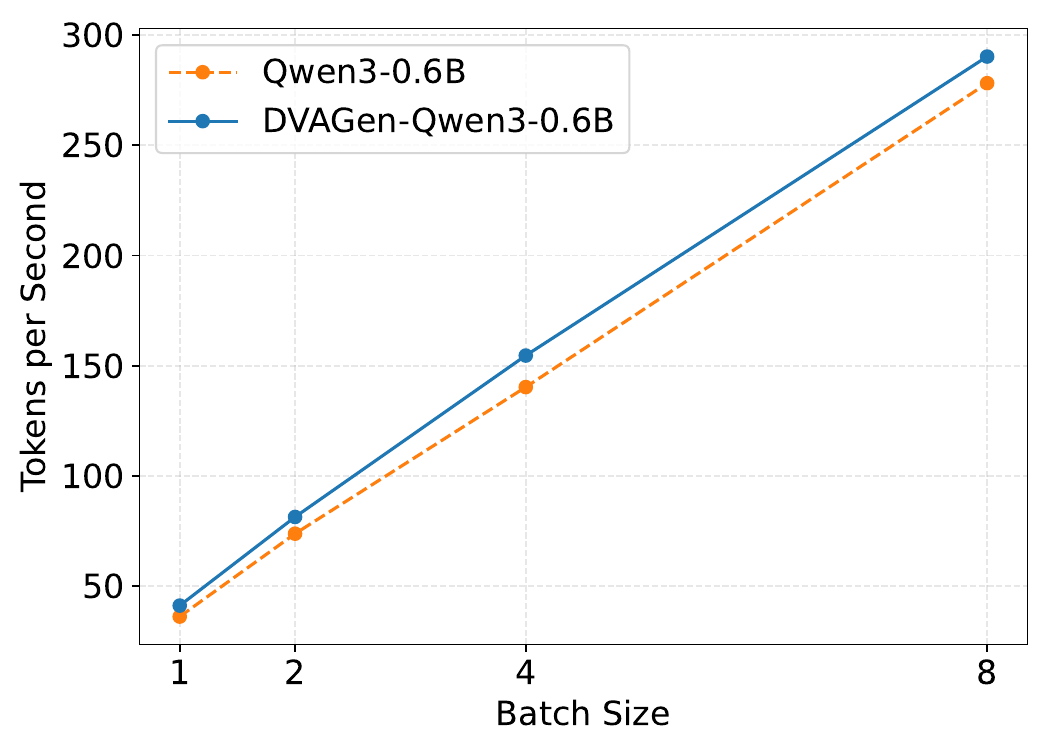}
        \caption{}
    \end{subfigure}
    \hfill
    \begin{subfigure}[b]{0.48\linewidth}
        \centering
        \includegraphics[width=\linewidth]{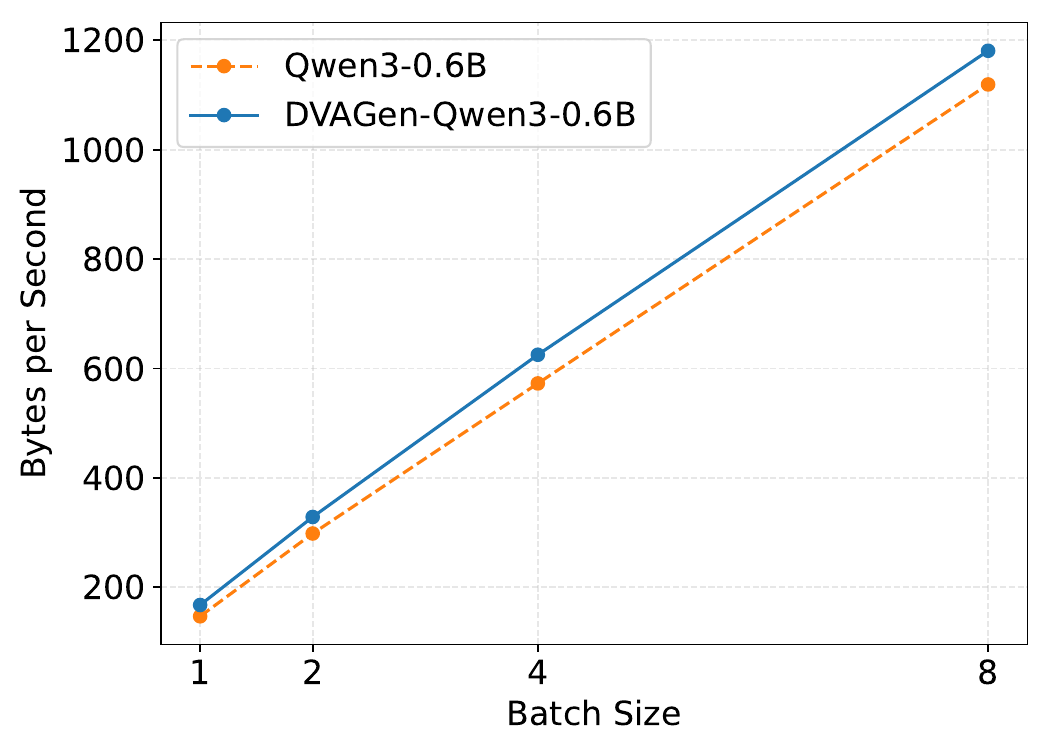}
        \caption{}
    \end{subfigure}
    \caption{Evaluation of performance in generating tokens and UTF-8 encoded bytes per second.}
    \label{fig:eval-inference-performance}
\end{figure}

\paragraph{Results}
The evaluation results are presented in Figure~\ref{fig:eval-inference-performance}, revealing two main findings:

\textbf{(1) Compared to the base model, \myframework achieves higher inference speed despite having more parameters.}
\myframework-Qwen3-0.6B employs two Qwen3-0.6B models, serving as the language model and phrase encoder, respectively, resulting in a model twice the size of the base model. However, due to phrases typically consisting of multiple tokens and containing more UTF-8 bytes, \myframework achieves faster inference than the base model under the same settings.

\textbf{(2) \myframework supports batch inference and consistently maintains high inference speed as the batch size increases.}
The implementation of batch inference improves generation efficiency by approximately \(7\times\) compared to processing one input at a time, as in the original implementations \cite{lan2023copy, liu2024generation}. Furthermore, the consistent inference performance with larger batch sizes demonstrates that \myframework effectively scales with workload, increases throughput, and optimizes GPU memory utilization without incurring additional latency.

\subsection{Retrieval Performance}
\label{sec:exp-retrieval}

\paragraph{Setup}
In addition to the inference performance (excluding the retrieval process) discussed in Section~\ref{sec:exp-inference}, we also conduct experiments to compare retrieval latency during inference. The experimental setup is similar to the previous one, except that a single NVIDIA RTX 3090 GPU is used. Since many real-time applications require longer contexts and the \retriever requires substantial GPU memory during inference, we also evaluate its retrieval performance on CPU devices. This experiment is conducted on a server equipped with an Intel(R) Xeon(R) Silver 4214R CPU @ 2.40 GHz and 512 GB of RAM.

\begin{figure}[!htb]
    \centering
    \begin{subfigure}[b]{0.48\linewidth}
        \centering
        \includegraphics[width=\linewidth]{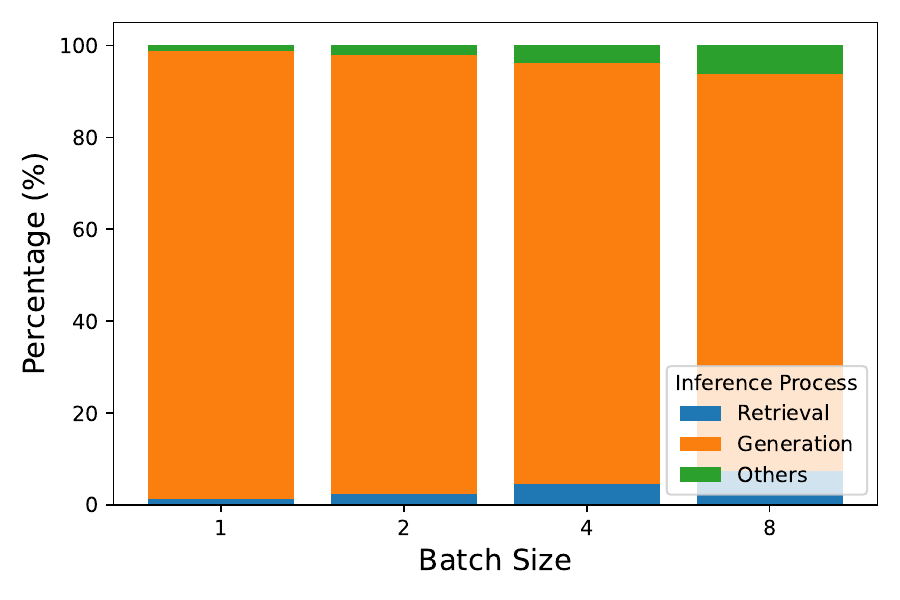}
        \caption{}
    \end{subfigure}
    \hfill
    \begin{subfigure}[b]{0.48\linewidth}
        \centering
        \includegraphics[width=\linewidth]{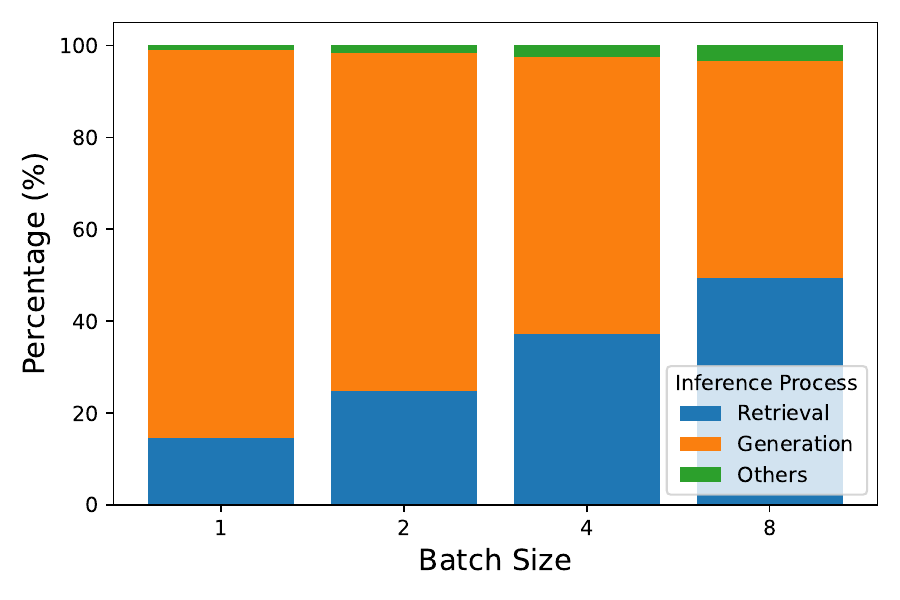}
        \caption{}
    \end{subfigure}
    \caption{Evaluation of retrieval performance: (a) retrieval performed on a GPU device, (b) retrieval performed on a CPU device.}
    \label{fig:eval-retrieval-performance}
\end{figure}

\paragraph{Results}
The comparisons of the proportions of different stages (primarily retrieval and generation) during inference are shown in Figure~\ref{fig:eval-retrieval-performance}. The key observations are as follows:

\textbf{(1) When a GPU is utilized for inference, the generation stage consistently dominates the overall inference time, substantially exceeding the time spent on retrieval.}
As discussed in Section~\ref{sec:infer}, \myframework retrieves relevant supporting documents on the GPU by default, with the retrieval process accounting for approximately 7\% of the inference time when the batch size is set to 8. In addition, since retrieval within a batch is performed sequentially due to the independence of samples, retrieval time increases with larger batch sizes.

\textbf{(2) Compared with GPU-based retrieval, CPU-based retrieval requires substantially more time.}
When the batch size is set to 8, CPU-based retrieval accounts for approximately half of the total inference time, slightly exceeding the duration of the generation stage (performed on a GPU device). This observation suggests that when using CPU-based retrieval to save GPU memory, the trade-off between batch size and inference latency should be carefully considered.

\section{Conclusion}
\label{sec:conclusion}
In this work, we present a fully open-source, modular framework that unifies training, evaluation, and visualization for dynamic vocabulary-augmented applications. The framework enables seamless customization of individual components and is the first to provide both CLI and WebUI tools for generation analysis. By supporting plug-and-play integration of existing open-source LLMs, we demonstrate the effectiveness of dynamic vocabulary methods in enhancing language modeling performance. Furthermore, with support for batch inference, our framework significantly improves inference throughput, offering a practical and scalable solution for deploying efficient LLM-based systems.


\section*{Acknowledgments}
The authors wish to thank all reviewers for their
helpful comments and suggestions. The corresponding author
is Yuanbin Wu.
This research was (partially) supported by National Key R\&D Program of China (2024YFC3308103).

\bibliography{reference}




\end{document}